\documentclass[12pt]{article}

\usepackage{subcaption}

\usepackage{float}

\usepackage{multirow}

\usepackage{tikz}

\usetikzlibrary{calc,backgrounds,arrows,matrix}

\usepackage{enumerate}
\usepackage{listings}

\usepackage{color}
\definecolor{lightblue}{rgb}{0,0.2,0.5}

\usepackage[colorlinks=true, urlcolor=blue,linkcolor=blue, citecolor=lightblue]{hyperref}

\usepackage{amssymb,amsmath}

\usepackage{graphicx}

\usepackage{color}
\usepackage{mathrsfs}

\usepackage{mathtools}
\usepackage{accents}

\DeclareMathAlphabet{\eufrak}{U}{}{}{}

\SetMathAlphabet\eufrak{normal}{U}{euf}{m}{n}
\SetMathAlphabet\eufrak{bold}{U}{euf}{b}{n}

\allowdisplaybreaks

 \def\qu{{\mathord{\mathbb Z}}}

 \def\sZZ{{\rm Z\kern-.45em{}Z}}

 \def\sQQ{{\kern 0.27em \vrule height1.45ex width0.03em depth0em
           \kern-0.30em \rm Q}}
 \def\qu{{\mathchoice
         {\sQQ}
         {\sQQ}
   {\kern 0.225em \vrule height1.05ex width0.025em depth0em \kern-0.25em \rm Q}
   {\kern 0.180em \vrule height0.78ex width0.020em depth0em \kern-0.20em \rm Q}
         }}
 \def\sGG{{\kern 0.27em \vrule height1.45ex width0.03em depth0em
           \kern-0.30em \rm G}}
 \def\gg{{\mathchoice
         {\sGG}
         {\sGG}
   {\kern 0.225em \vrule height1.05ex width0.025em depth0em \kern-0.25em \rm G}
   {\kern 0.180em \vrule height0.78ex width0.020em depth0em \kern-0.20em \rm G}
         }}

 \newtheorem{prop}{Proposition}[section]

 \newtheorem{corollary}[prop]{Corollary}

 \newtheorem{remark}[prop]{Remark}
 
 \newtheorem{assumption}{Assumption}

\numberwithin{equation}{section}

\newcommand{\norm}[1]{\lVert#1\rVert}
 \def\P{{\mathord{\mathbb P}}}

 \newcounter{hyp}
\newenvironment{Proof}{\removelastskip\par\medskip \noindent{\em Proof.} \rm}{\penalty-20\null\hfill$\square$\par\medbreak}

\def\bprf{\begin{Proof}}
\def\nprf{\end{Proof}}
\def\bdes{\begin{description}}
\def\ndes{\end{description}}

\newtheorem{thm}{Theorem}[section]

\def\bdef{\begin{defn}}
\def\ndef{\end{defn}}
\def\bthm{\begin{thm}}
\def\nthm{\end{thm}}
\def\bprop{\begin{prop}}
\def\nprop{\end{prop}}
\def\brmk{\begin{remark}}
\def\nrmk{\end{remark}}
\def\bexa{\begin{exa}}
\def\nexa{\end{exa}}
\def\blem{\begin{lem}}
\def\nlem{\end{lem}}
\def\bcor{\begin{cor}}
\def\ncor{\end{cor}}
\def\bexe{\begin{exe}}
\def\nexe{\end{exe}}

\newcommand{\E}{\mathbb{E}}

\newcommand{\real}{\mathbb{R}}

\newcommand{\Supp}{{\rm Supp \ \! \! }}

\def\og{\leavevmode\raise.3ex
     \hbox{$\scriptscriptstyle\langle\!\langle$~}}
\def\fg{\leavevmode\raise.3ex
     \hbox{~$\!\scriptscriptstyle\,\rangle\!\rangle$}~}

\title{\Huge
 Generalization error bounds for two-layer neural networks
 with Lipschitz loss function
}

\author{
 Jiang Yu Nguwi\footnote{\href{mailto:nguw0003@e.ntu.edu.sg}{nguw0003@e.ntu.edu.sg}
 }
 \qquad Nicolas Privault\footnote{
\href{mailto:nprivault@ntu.edu.sg}{nprivault@ntu.edu.sg}
 }
 \\
  \small
Division of Mathematical Sciences
\\
\small
School of Physical and Mathematical Sciences
\\
\small
Nanyang Technological University
\\
\small
21 Nanyang Link, Singapore 637371
}

\allowdisplaybreaks

\usepackage{empheq}
\usepackage{bm}

\usepackage{bbm}

\makeatletter
\newcommand*\rel@kern[1]{\kern#1\dimexpr\macc@kerna}
\newcommand*\widebar[1]{%
  \begingroup
  \def\mathaccent##1##2{%
    \rel@kern{0.8}%
    \overline{\rel@kern{-0.8}\macc@nucleus\rel@kern{0.2}}%
    \rel@kern{-0.2}%
  }%
  \macc@depth\@ne
  \let\math@bgroup\@empty \let\math@egroup\macc@set@skewchar
  \mathsurround\z@ \frozen@everymath{\mathgroup\macc@group\relax}%
  \macc@set@skewchar\relax
  \let\mathaccentV\macc@nested@a
  \macc@nested@a\relax111{#1}%
  \endgroup
}
\makeatother

\begin{document}
\maketitle

\baselineskip0.6cm

\vspace{-0.6cm}

\begin{abstract}
 We derive generalization error bounds
 for the training of two-layer neural networks
 without assuming boundedness of the loss function, 
 using Wasserstein distance estimates on the discrepancy 
 between a probability distribution and its 
 associated empirical measure, together with moment bounds
 for the associated stochastic gradient method. 
 In the case of independent test data, we obtain
 a dimension-free rate of order $O\big(n^{-1/2} \big)$ 
 on the $n$-sample generalization error, whereas without independence assumption,
 we derive a bound of order $O\big(n^{-1 / ( d_{\rm in}+d_{\rm out} )} \big)$, 
 where $d_{\rm in}$, $d_{\rm out}$ denote input and output dimensions. 
 Our bounds and their coefficients 
 can be explicitly computed prior to the training of the model,  
 and are confirmed by numerical simulations.
\end{abstract}



\baselineskip0.7cm

\parskip-0.1cm

\section{Introduction}
The study of two-layer neural networks using stochastic gradient
descent and their approximation
guarantees has attracted considerable attention in recent years,
see e.g. \cite{mei2019meanfield} and references therein
 for a review using mean-field theory, 
 \cite{neu2021information}
 for the use of information-theoretic generalization bounds,
 or \cite{park2022generalization}
 for covering arguments. 

 \medskip
 
The aim of the present paper is to propose generalization bounds
for two-layer neural networks without assuming the boundedness
of loss and activation functions, using
bounds established in \cite{fournier2015rate}
on the Wasserstein  
 between a probability distribution and its 
 associated empirical measure. 
 
 \medskip
 
Let 
$$
 Z: = (Z_i)_{1 \leq i \leq n}
 = (X_i, Y_i)_{1 \leq i \leq n} \overset{\text{i.i.d.}}{\sim} \rho
 $$
 be a set of $n\geq 1$ independent data samples $( X_i , Y_i)
 \in \real^{d_{\rm in}} \times \real^{d_{\rm out}}$, identically
 distributed according to a probability distribution $\rho$ %
on $\real^{d_{\rm in}}\times \real^{d_{\rm out}}$ which is not accessible in practice.
 Consider
 $$
 l: \real^{d_{\rm out}} \times \real^{d_{\rm out}} \to \real
 $$
 a loss function, and 
 a two-layer neural network function
$$
 f( \cdot , v, w ) : \real^{d_{\rm in}} \to \real^{d_{\rm out}} 
 $$
 parameterized by matrices 
 $v \in \real^{d_{\rm in} \times d}$, 
 $w \in \real^{d \times d_{\rm out}}$,
 $d, d_{\rm in}, d_{\rm out}\geq 1$. 

 \medskip

 In this context, given the output 
 $(V(t),W(t)) \in \real^{d_{\rm in} \times d} \times \real^{d \times d_{\rm out}}$ 
 of a Stochastic Gradient Method (SGM) 
 trained on data sets $Z^{(t)}=\big(X_i^{(t)},Y_i^{(t)}\big)_{1 \leq i \leq n}$ 
 of independent samples identically distributed according to $\rho$ 
 at epochs $t=0,1,\ldots , T$, %
 we consider the generalization error defined as the difference 
\begin{align} 
  \label{eq:gen error}
  & \varepsilon_{\rm gen}(n, V(T), W(T))
  \\
  \nonumber
  & \qquad \quad
  :=
  \int_{\real^{d_{\rm in}}\times \real^{d_{\rm out}}} l(f (x ,  V(T), W(T) ), y) \rho(dx, dy)
  -
  \frac{1}{n} \sum\limits_{i=1}^n l(f ( X_i , V(T), W(T) ), Y_i) 
 \quad 
\end{align}
 between the expected loss of the neural network under 
 the true data distribution $\rho(dx, dy)$
 and its average loss on the training dataset
 $(X_i, Y_i)_{1 \leq i \leq n}$, which measures how well the model 
 generalizes to the unknown underlying distribution $\rho$.
 
\medskip

 In the case of a uniformly bounded loss function, 
 a $0$-$1$ error bound of order $O\big(n^{-1/2}\big)$
 on the expected generalization error has been obtained in 
 \cite{cao2019generalization}. 
 Related bounds have been derived in
\cite{hardt2016train}, 
\cite{raginsky2017non}, 
\cite{mou2018generalization}, 
\cite{pensia2018generalization},  
and in \cite{zhang2022stability} 
using a stability approach, by 
further assuming the boundedness of the gradients
$\nabla_v l(f ( x , v, w ), y)$ and
$\nabla_w l(f ( x , v, w ), y)$, or the $\beta$-smoothness
property, or under a uniform boundedness assumption on the loss function 
\cite{wang2025generalization}.
 See also \cite{allen2019learning} for the case of three-layer networks,
 and \cite{sncohen} for the derivation of
 $O\big(n^{-1}\big)$ error bounds
 using calculus on the space of measures.

\medskip

In this paper, we aim at bounding
$\big| \varepsilon_{\rm gen}(n, V(T), W(T)) \big|$
 in a context where the loss function 
 $l(y_1,y_2)$ may not be bounded,
 by relaxing the boundedness of $l(f ( x , v, w ), y)$ or its gradients 
 using a Lipschitz condition  
 which is satisfied by loss functions such as 
 the mean absolute error or the Huber loss function.
 We also require a ${\cal C}^1$ Lipschitz condition on the activation
 function of the neural network,
 which is satisfied by e.g. the %
 softplus, tanh, and sigmoid functions.

 \medskip

 In Proposition~\ref{main prop}, we start by deriving 
 moment bounds of the SGM output $(V(T),W(T))$
 for two-layer neural networks. 
 Next, using a testing data set
$$
Z = (Z_i)_{1 \leq i \leq n}
= (X_i, Y_i)_{1 \leq i \leq n}
\overset{\text{i.i.d.}}{\sim} \rho
$$
independent of $\big(Z^{(t)}\big)_{0 \leq t \leq T}$ 
 and Proposition~\ref{main prop}, 
 we derive an error bound of dimension-free
 order $O\big(n^{-1 / 2} \big)$
 on the $L^1$ norm $\E \big[
   \big| \varepsilon_{\rm gen}(n, V(T), W(T))\big|
   \big]$,
 see Proposition~\ref{gen error prop independent},
 and related deviation inequalities 
 in Proposition~\ref{gen error prop independent-2}. 
  
 \medskip
  
 In Proposition~\ref{gen error prop},
 without the above independence assumption, 
 we apply the Wasserstein distance bounds of \cite{fournier2015rate}
 and Proposition~\ref{main prop} 
 to derive generalization error bounds of order
 $O\big(n^{-1 / ( d_{\rm in}+d_{\rm out} )} \big)$ 
 on the expectation and deviation probability of
 $\big|\varepsilon_{\rm gen}(n, V(T), W(T))\big|$.
 Related dimension-dependent phenomena have been observed
 in e.g.\ \cite{finlay2018lipschitz}
 when no boundedness is assumed on the loss function and its gradient.
 In Proposition~\ref{main prop full} 
 we also derive $L^p$ bounds on the Lipschitz constant 
 of regularized loss functions, 
 with concentration inequalities presented in Corollary~\ref{fjkl132}.

 \medskip

 Unlike in e.g. 
 \cite{xu2017information}, \cite{lopez2018generalization},
 \cite{arora2019fine}, \cite{wang2019information},
 where the bounds rely on quantities that may not be available in practice,
 all constants appearing in our bounds
 can be explicitly computed without actually training the network.
In contrast, the bounds derived by 
\cite{neyshabur2015norm}, 
\cite{neyshabur2017pac}, 
\cite{dziugaite2017computing}, 
\cite{neyshabur2018role},  
\cite{cao2019generalization} 
rely on some properties of a trained network
that are unknown before the training.

 \medskip

This paper is organized as follows.
In Section~\ref{sec:prelim},
we introduce the SGM dynamics, its generalization error,
and the Wasserstein distance bounds of \cite{fournier2015rate}. 
In Section~\ref{sec:optimize last layer}, 
we derive moment bounds for the SGM dynamics, 
see Proposition~\ref{main prop}. 
 In Section~\ref{sec:application to gen error independent}
 we obtain $L^1$ error bounds in the case where
 the testing set is independent of 
 the training data sequence used for SGM updates, see
 Proposition~\ref{gen error prop independent}
 and the deviation inequalities 
 of Proposition~\ref{gen error prop independent-2}.
 Generalization error bounds
 without independence assumption are obtained in 
 Section~\ref{sec:application to gen error}, see 
 Proposition~\ref{gen error prop}. 
 This is followed by bounds and concentration inequalities
 for the Lipschitz constant of the loss function
 in Proposition~\ref{main prop full}
 and Corollary~\ref{fjkl132}.
 Numerical confirmations 
 are presented in Section~\ref{sec:numerical results}.

\section{Preliminaries and notation}
\label{sec:prelim}
For $x = (x_1, \ldots, x_d)^\top$ a vector in $\real^d$ 
we use the Euclidean norm defined by
$|x| = \sqrt{x_1^2+\cdots + x_d^2}$.
For $v \in \real^{m \times n}$ a matrix, 
we let
$$
\norm{v}_F: =
\sqrt{
  \sum_{i=1}^m 
\sum_{j=1}^n 
|v_{i,j}|^2}
$$ 
 denote %
 the matrix Frobenius %
 norm of $v$, and 
 we let $\Supp (\mu )$ denote the support of any
 probability measure~$\mu$. 
\subsubsection*{Wasserstein distance}
Let $p \geq 1$.
 For $\mu, \nu$ in the space
 $\mathcal{P}_p \big( \real^{d_{\rm in} + d_{\rm out}} \big)$
 of probability measures on $\real^{d_{\rm in} + d_{\rm out}}$
 with finite $p$-moment,
 the Wasserstein-$p$ distance between $\mu$ and $\nu$ is defined as
\begin{equation*}
\mathcal{W}_p(\mu, \nu) :=
\inf_{
  \pi \text{  coupling }
  \atop
  \text{ of $\mu$ and $\nu$}
}
\bigg(\int_{( \real^{d_{\rm in} + d_{\rm out}} )^2 %
} |z_1-z_2|^p \pi(dz_1, dz_2) \bigg)^{1/p}. 
\end{equation*}
 Recall that by \cite{kantorovich1958},
 for any $\mu, \nu \in \mathcal{P}_1\left(\real^{d_{\rm in} + d_{\rm out}}\right)$
 we have
\begin{equation}
    \label{kantorovich-rubinstein}
        \mathcal{W}_1(\mu, \nu)
        = \sup_{
          h \text{ is $1$-Lipschitz}
          \atop \text{on } \real^{d_{\rm in} + d_{\rm out}}
          }
        \left(
        \int_{\real^{d_{\rm in} + d_{\rm out}}} h d\mu - \int_{\real^{d_{\rm in} + d_{\rm out}}} h d\nu
        \right), 
\end{equation}
 and from \cite{fournier2015rate} we have the following
 proposition, where $\delta_x$ represents the Dirac delta at the point~$x$.
\begin{prop}
\label{fg}
 \cite[Theorems 1 and 2]{fournier2015rate}. 
 Suppose that $d_{\rm in} + d_{\rm out} \geq 5$, and let 
$$
\tilde{\rho}_n := \frac{1}{n} \sum\limits_{i=1}^n \delta_{(X_i, Y_i)}(dz)
$$
 denote the empirical measure associated to the
 sequence $(Z_i)_{1 \leq i \leq n} = (X_i, Y_i)_{1 \leq i \leq n}$.
 \begin{enumerate}[a)]
 \item
   We have the Wasserstein bound 
\begin{equation}
    \label{fournier guillin}
    \E \left[ \mathcal{W}^2_F (\rho, \tilde{\rho}_n) \right]
    \leq C n^{ - 2 / ( d_{\rm in}+d_{\rm out} ) }. 
\end{equation}
\item For any $\zeta \in (0,1)$, we have the concentration inequality 
\begin{equation}
    \label{fournier guillin_2}
    \P \left( \mathcal{W}_1(\rho, \tilde{\rho}_n)
    \leq \left(
    \frac{Cn}{\log(C / \zeta)}
    \right)^{- 1 / ( d_{\rm in}+d_{\rm out} ) }
    \right)
    \geq 1 - \zeta,
\end{equation}
where $C>0$ is a constant independent of $\zeta \in (0,1)$.
 \end{enumerate}
\end{prop}
\subsubsection*{SGM dynamics} %
For $\lambda > 0$, let $\ell_\lambda$ denote the
 loss function 
 \begin{equation}
   \label{jkl131} 
    \ell_\lambda ( x, y , v, w) := l(x, y)
    + \frac{\lambda}{2} \left(\norm{v}^2_F
     + \norm{w}^2_F \right),
\end{equation}
 where $\norm{v}_F$, $\norm{w}_F$ 
 are the Frobenius norms of 
 $v \in \real^{d_{\rm in} \times d}$, 
 $w \in \real^{d \times d_{\rm out}}$,  
 and $\lambda > 0$ is a regularization parameter. 
 Given a neural network function $f ( \cdot , v, w )
:\real^{d_{\rm in}} \to \real^{d_{\rm out} }$  parameterized by
 $v \in \real^{d_{\rm in} \times d}$ and $w \in \real^{d \times d_{\rm out}}$, 
 we aim at finding the infimum
 \begin{equation}
   \label{fjklda} 
  \inf\limits_{
    (v, w)\in \real^{d_{\rm in} \times d} \times \real^{d \times d_{\rm out}}
  } \mathcal{L}(v, w), 
\end{equation}
 where 
 \begin{equation}
   \label{lwv} 
\mathcal{L}(v, w) := \int_{\real^{d_{\rm in}}\times \real^{d_{\rm out}}} \ell_\lambda ( f ( x , v, w ), y,v, w) \rho(dx, dy). 
\end{equation} 
 As we do not have the access to the actual data distribution $\rho$,
 given 
$$
Z^{(t)}=\big(X_i^{(t)},Y_i^{(t)}\big)_{i=1,\ldots , k}
$$
 a set of independent data samples $\big( X_i^{(t)} , Y_i^{(t)}\big)
 \in \real^{d_{\rm in}} \times \real^{d_{\rm out}}$ of batch size $k\geq 1$, 
 identically distributed according to $\rho$
 at times $t\geq 0$, 
 we approximate \eqref{fjklda} by minimization of 
\begin{equation}
\nonumber %
    \inf\limits_{
      (v, w)\in \real^{d_{\rm in} \times d} \times \real^{d \times d_{\rm out}}
      } \mathcal{L}_k \big( Z^{(t)} , v, w\big) 
\end{equation}
 where 
 \begin{equation}
   \label{fjkl1} 
\mathcal{L}_k \big( Z^{(t)} , v, w\big)
:= \frac{1}{k} \sum\limits_{i=1}^k
\ell_\lambda \big( f \big( X_i^{(t)} , v, w \big), Y_i^{(t)} , v, w\big). 
\end{equation} 
 For this, we use the sequence $(V(t),W(t))_{0\leq t \leq T}$ defined
 by the Stochastic Gradient Method (SGM), through the dynamics
\begin{equation}
    \label{sgm dynamics}
    \left\{
    \begin{array}{l}
      V(t+1) = V(t)
      - \eta_V (t) \nabla_v \mathcal{L}_k \big( Z^{(t)} , V(t), W(t) \big),
      \medskip
      \\ 
    W(t+1) = W(t)
    - \eta_W (t) \nabla_w \mathcal{L}_k \big( Z^{(t)} ,  V(t), W(t) \big),
    \quad t = 0,1,\ldots , T-1, 
    \end{array}
    \right.
    \end{equation}
 where $(\eta_V (t))_{0 \leq t < T}$ and
 $(\eta_W (t))_{0 \leq t < T}$ denote (positive) learning rate
 sequences.
 
\medskip

 We will work under the following conditions.
\begin{assumption}
\label{basic assumptions}
We assume that
 \begin{itemize}
\item $\Supp (\rho) \subset \{ z = (x,y) \in
    \real^{d_{\rm in}}\times \real^{d_{\rm out}} \ : \
    \max (|x|, |y|) \leq 1 \}$,
\item
 the function
 $l: \real^{d_{\rm out}} \times \real^{d_{\rm out}} \to \real$ 
 is ${\cal C}^1$, $1$-Lipschitz, and satisfies 
 $$
 l(y, y) = 0,
 \quad
 y \in \real^{d_{\rm out}},
 \ |y|\leq 1,
$$ 
\item 
 $f ( x , v, w )$ is a two-layer neural network
of the form
\begin{equation}
  \label{dkjlda}
  f ( x , v, w ) = w^\top \sigma \big(v^\top x\big),
  \quad x\in \real^{d_{\rm in}},
\end{equation}
 where $\sigma: \real \to \real$ is a 
 ${\cal C}^1$ and $1$-Lipschitz activation function such that $\sigma(0) = 0$,
 which is applied componentwise to $v^\top x\in \real^d$,
\item
  the SGD dynamics satisfies the learning rate conditions 
  $$
  0 \leq \eta_W (t) \leq \eta_V (t) \leq \frac{1}{\lambda},
  \quad 
  0 \leq t < T.
  $$ 
\item The entries of the matrices
 $V(0) \in \real^{d_{\rm in} \times d}$
 and
 $W(0) \in \real^{d \times d_{\rm out}}$ 
 are initialized via He initialization, using 
 independent centered Gaussian samples
 with variance $\kappa/d$ (resp. $\kappa/d_{\rm out}$),
 with $\kappa = 2$, %
 see \cite{he2015delving}.
\end{itemize}
\end{assumption}
 We note that by taking $K>0$, Assumption~\ref{basic assumptions} can be relaxed
by only assuming that $\max (|x|,|y|) \leq K$
for all $(x,y) \in \Supp (\rho)$,
and that the function $l$ and activation function $\sigma$ are $K$-Lipschitz. 
\section{SGM moment bounds} %
\label{sec:optimize last layer}
 In this section,
 we control the spectral norms of $V(T)$ and $W(T)$ in the SGM dynamics
 \eqref{sgm dynamics}.
 We let $p !!$ denote the double factorial of $p \geq 0$.
\begin{prop}
\label{main prop}
    Moment bounds.
    Suppose that Assumption~\ref{basic assumptions} holds. 
\begin{enumerate}[a)]
\item If $W(t)$ remains frozen at
 $W(t):=w \in \real^{d \times d_{\rm out}}$, 
 i.e. $\eta_W (t) = 0$, $0 \leq t < T$, 
 then for all $p \geq 1$ we have  
\begin{equation} 
  \label{aa1}
  \E \left[ \norm{V(T)}^p_F \right]
     \leq 
        (p - 1)!! 2^{p-1} (\kappa d_{\rm in})^{p/2}
    \left(\prod\limits_{t = 0}^{T - 1} (1 - \eta_V (t) \lambda)\right)^p
    +
    2^{p-1} \frac{\norm{w}_F^p}{\lambda^p}
    \left(1 - \prod\limits_{t = 0}^{T - 1} (1 - \eta_V (t) \lambda)\right)^p. 
\end{equation} 
\item
  If $\eta_W (t) = \eta_V (t):=\eta (t)$, $0 \leq t < T$, 
  then for any $p\geq 1$ we have
  \begin{equation}
    \label{aa2} 
        \E \big[ \norm{V(T)}^p_F \norm{W(T)}^p_F \big]
        \leq 
                \frac{\kappa^d}{2} 
                (2p - 1)!!
                \big(
 d_{\rm in}^p
 +
 d_{\rm out}^p
 \big) 
            \prod\limits_{t = 0}^{T - 1}
            (1 - \eta (t) \lambda + \eta (t))^{2p}. 
\end{equation}
\end{enumerate}
\end{prop}
\begin{Proof}
 From \eqref{dkjlda}, the regularized loss function
 \eqref{jkl131} satisfies 
  \begin{equation*}
    \ell_\lambda ( f(x,v,w), y , v, w) = l\big(w^\top \sigma\big(v^\top x\big), y\big)
    + \frac{\lambda}{2} \left(\norm{v}^2_F
     + \norm{w}^2_F \right),
\end{equation*}
 hence 
\begin{align}
\nonumber %
   \nabla_v \ell_\lambda ( f(x,v,w), y , v, w)
     = x \big( ( \nabla_{y_1} l ) \big(w^\top \sigma\big(v^\top x\big), y\big)\big)^\top w^\top \Sigma + \lambda v,
\end{align}
where
$$
\Sigma := {\rm Diag}\big(\sigma'\big(v^\top x\big)\big)
$$
 is the square diagonal matrix
 with $\sigma'\big(v^\top x\big)$ as diagonal entries, and 
 the derivative $\sigma'$ of the activation function
 $\sigma$ is applied componentwise to $v^\top x \in \real^d$. 
 Therefore, from \eqref{fjkl1} we have
\begin{align*} 
  \nabla_v \mathcal{L}_k \big( Z^{(t)} , v, w \big)
 & = \frac{1}{k} \sum\limits_{i=1}^k
  \nabla_v
  \ell_\lambda \big( f \big( X_i^{(t)} , v, w \big), Y_i^{(t)} , v, w \big) 
 \\
 & = \frac{1}{k} \sum\limits_{i=1}^k
 \big(
 X_i^{(t)} \big(\nabla_{y_1} l\big(w^\top \sigma \big(v^\top X_i^{(t)}\big),
 Y_i^{(t)}\big)\big)^\top
 w^\top \Sigma + \lambda v
 \big)
 , 
\end{align*} 
 and \eqref{sgm dynamics} yields 
\begin{align*} 
      V(t+1) & = V(t)
      - \eta_V (t) \nabla_v \mathcal{L}_k \big( Z^{(t)} , V(t), W(t) \big)
      \\
      & =
      \big( 1 - \eta_V (t) \lambda \big) V(t)
      - \frac{\eta_V (t)}{k} \sum\limits_{i=1}^k
 X_i^{(t)} \big(\nabla_{y_1} l\big(W(t)^\top \sigma \big(v^\top X_i^{(t)}\big),
 Y_i^{(t)}\big)\big)^\top
 W(t)^\top \Sigma 
       ,
\end{align*} 
hence from the bound 
\begin{equation}
    \label{all bounds}
    \max
    (|\nabla_{y_1} l(y_1,y_2)|, |\nabla_{y_2} l(y_1,y_2)| 
    ) \leq 1,
    \quad
    y_1,y_2 \in \real^{d_{\rm out}}, 
\end{equation}
 we have
    \begin{align}
  &    \nonumber 
      \norm{V(t+1)}_F
    =
\norm{ V(t)
      - \eta_V (t) \nabla_v \mathcal{L}_k \big( Z^{(t)} , V(t), W(t) \big)}_F
\\
\nonumber 
& \quad
 \leq
(1 - \eta_V (t) \lambda) \norm{V(t)}_F +
 \frac{\eta_V (t)}{k} \sum\limits_{i=1}^k
 \big\Vert
 X_i^{(t)} \big(\nabla_{y_1} l\big(W(t)^\top \sigma \big(v^\top X_i^{(t)}\big),
 Y_i^{(t)}\big)\big)^\top
 W(t)^\top \Sigma
 \big\Vert_F
\\
\label{jkls1}
& \quad
 \leq
        (1 - \eta_V (t) \lambda) \norm{V(t)}_F + \eta_V (t) \norm{W(t)}_F.
    \end{align}
 Next, from the relation 
$$ 
    \nabla_w \ell_\lambda ( f(x,v,w), y , v, w)
    = \sigma\big(v^\top x\big)
    \big(
    (\nabla_{y_1} l ) \big(w^\top \sigma\big(v^\top x\big), y\big)\big)^\top + \lambda w 
$$ 
 and \eqref{fjkl1}, we have 
\begin{align*} 
 \nabla_w \mathcal{L}_k \big( Z^{(t)} , V(t), W(t) \big)
 & = \frac{1}{k} \sum\limits_{i=1}^k
 \nabla_w \big( \ell_\lambda \big( f \big( X_i^{(t)} , V(t), W(t) \big), Y_i^{(t)} , V(t), W(t)\big) \big)
 \\
 & = \frac{1}{k} \sum\limits_{i=1}^k
 \big(
 \sigma
 \big(V(t)^\top X_i^{(t)}\big)
 \big( (\nabla_{y_1} l) \big(W(t)^\top \sigma\big(V(t)^\top X_i^{(t)}\big), Y_i^{(t)}\big)\big)^\top
 + \lambda W(t)
 \big)
 , 
\end{align*} 
 and \eqref{sgm dynamics} yields 
\begin{align*} 
      W(t+1) & = W(t)
    - \eta_W (t) \nabla_w \mathcal{L}_k \big( Z^{(t)} ,  V(t), W(t) \big)
      \\
     & =
      \big( 1 - \eta_W (t) \lambda \big) W(t)
      - \frac{\eta_W (t)}{k} \sum\limits_{i=1}^k
 \sigma
 \big(V(t)^\top X_i^{(t)}\big)
 \big(\nabla_{y_1} l\big(W(t)^\top \sigma\big(V(t)^\top X_i^{(t)}\big), Y_i^{(t)}\big)\big)^\top
       ,
\end{align*} 
 hence from \eqref{all bounds} we have
    \begin{align}
      \nonumber
 &     \norm{W(t+1)}_F
      =
\norm{ W(t)
      - \eta_W (t) \nabla_w \mathcal{L}_k ( Z^{(t)} , V(t), W(t) )}_F
\\
\nonumber
&
\quad
\leq
(1 - \eta_W (t) \lambda) \norm{W(t)}_F 
   + \frac{\eta_W (t)}{k} \sum\limits_{i=1}^k
   \big\Vert
   \sigma
 \big(V(t)^\top X_i^{(t)}\big)
 \big(\nabla_{y_1} l\big(W(t)^\top \sigma\big(V(t)^\top X_i^{(t)}\big), Y_i^{(t)}\big)\big)^\top
\big\Vert_F
\\
\nonumber
&
\quad
\leq
        (1 - \eta_W (t) \lambda) \norm{W(t)}_F + \eta_W (t) \norm{V(t)}_F
  \\
  \label{fkjl143}
  &
  \quad
\leq        (1 - \eta_W (t) \lambda) \norm{W(t)}_F + \eta_V (t) \norm{V(t)}_F. 
\end{align} 
\noindent
$a)$ If $\eta_W (t) = 0$, $0 \leq t < T$,
 then from \eqref{jkls1} and \eqref{fkjl143}
 we have
 \begin{align}
        \nonumber
        \norm{V(T)}_F
        & \leq \norm{V(0)}_F
        \left(\prod\limits_{t = 0}^{T-1} (1 - \eta_V (t) \lambda)\right)
        + \norm{W(0)}_F
        \sum\limits_{t = 0}^{T-1} \eta_V (t)
        \prod\limits_{i = t + 1}^{T - 1} (1 - \eta_V (i) \lambda)
        \\
   \label{jklsd18} 
        & = \norm{V(0)}_F
        \left(\prod\limits_{t = 0}^{T-1} (1 - \eta_V (t) \lambda)\right)
        + \frac{\norm{w}_F}{\lambda}
        \left(1 - \prod\limits_{t = 0}^{T - 1} (1 - \eta_V (t) \lambda)\right). 
    \end{align}
    \noindent
 We conclude by taking expectations on both sides and
 noting that since the matrix
 $V(0) = (v_{i,j})_{(i,j) \in d_{\rm in} \times d} \in \real^{d_{\rm in} \times d}$
 has centered independent Gaussian entries with variance $\kappa / d$, we have 
 \begin{align}
\nonumber 
   \E \left[\norm{V(0)}^p_F\right]
   & =
   \E \bigg[
     \bigg(
     \sum_{i=1}^{d_{\rm in}} 
     \sum_{j=1}^d |v_{i,j}|^2\bigg)^{p/2}
       \bigg]
   \\
   \nonumber
   & \leq 
     ( d_{\rm in} d)^{p/2-1}
          \sum_{(i,j) \in d_{\rm in} \times d}      \E \left[|v_{i,j}|^p
     \right]
\\
   \label{jkls13-0} 
      & = 
(p - 1)!! (\kappa d_{\rm in})^{p/2},
\quad p\geq 1. 
\end{align} 
 
        \noindent
$b)$ 
        If $\eta_V (t) = \eta_W (t) := \eta (t)$, $0 \leq t < T$,
         then from \eqref{jkls1} and \eqref{fkjl143} we have
\begin{equation}
\label{b22} 
 \norm{V(T)}_F + \norm{W(T)}_F 
 \leq
 ( \norm{V(0)}_F
      + \norm{W(0)}_F ) 
            \prod\limits_{t = 0}^{T - 1}
            (1 - \eta (t) \lambda + \eta (t)), 
    \end{equation}
 hence
\begin{align*} 
 \E \big[ \norm{V(T)}^p_F \norm{W(T)}^p_F \big]
 & \leq 
 2^{-2p} \E \big[ ( \norm{V(T)}_F + \norm{W(T)}_F)^{2p} \big]
\\
 & \leq 
 2^{-2p}
 \E \big[ ( \norm{V(0)}_F
      + \norm{W(0)}_F )^{2p} \big] 
            \prod\limits_{t = 0}^{T - 1}
            (1 - \eta (t) \lambda + \eta (t))^{2p}
\\
 & \leq 
 \frac{1}{2} 
 \E \big[ \norm{V(0)}_F^{2p}
      + \norm{W(0)}_F ^{2p} \big] 
            \prod\limits_{t = 0}^{T - 1}
            (1 - \eta (t) \lambda + \eta (t))^{2p}
\\
 & \leq 
\kappa^d 
\frac{(2p - 1)!! }{2} 
 \big(d_{\rm in}^p
 +
 d_{\rm out}^p
 \big) 
            \prod\limits_{t = 0}^{T - 1}
            (1 - \eta (t) \lambda + \eta (t))^{2p}, 
\end{align*} 
 since, as in~\eqref{jkls13-0}, we have 
\begin{equation}
\nonumber %
\E \left[\norm{V(0)}^{2p}_F\right] \leq (2p - 1)!! (\kappa d_{\rm in} )^p 
\quad
\mbox{and}
\quad 
\E \left[\norm{W(0)}^{2p}_F\right] \leq (2p - 1)!! (\kappa d_{\rm out} )^p, 
\quad p\geq 1, 
\end{equation}
 see also %
 \cite{rudelson2010non}. 
\end{Proof}
 We note that the upper bounding constants in Proposition~\ref{main prop}
 and in subsequent results remain bounded as $T$ tends to infinity,
 provided that the sequence $(\eta_V (t))_{t\geq 0}$ is summable, i.e.
$$
 \sum_{t\geq 0} |\eta_V (t) | < \infty, 
$$ 
 which is the case in particular for schedules with polynomial
 time decay of the form $1/a^t$, $a>1$. 

 \medskip

  In the case of constant schedules
$\eta_W (t) = \eta_V (t)=\eta$, $0 \leq t < T$,
the bounds \eqref{aa1} and \eqref{aa2} become
\begin{equation}
  \label{aa1-1} 
     \E \left[ \norm{V(T)}^p_F \right]
     \leq 
        (p - 1)!! 2^{p-1} (\kappa d_{\rm in})^{p/2}
    (1 - \lambda \eta )^{pT} 
    +
    2^{p-1} \frac{\norm{w}_F^p}{\lambda^p}
    \big(1 - (1 - \eta \lambda)^T\big)^p
\end{equation}
    and
\begin{equation}
  \label{aa2-2} 
        \E \big[ \norm{V(T)}^p_F \norm{W(T)}^p_F \big]
        \leq 
                \frac{\kappa^d}{2} 
                (2p - 1)!!
                \big(
 d_{\rm in}^p
 +
 d_{\rm out}^p
 \big) 
 (1 + (1 - \lambda ) \eta )^{2pT}. 
\end{equation}
 In this case, \eqref{aa1-1} tends to
 $2^{p-1} {\norm{w}_F^p} / {\lambda^p}$
 while \eqref{aa2-2} explodes as $T$ tends to infinity. 
\section{%
Independent samples} %
\label{sec:application to gen error independent}
 In Proposition~\ref{gen error prop independent}
 we derive $L^1$ error bounds of dimension-free order for 
 the absolute generalization error 
 $\big| \varepsilon_{\rm gen}(n, V(T), W(T)) \big|$ 
 by assuming as in \cite{kawaguchi2017generalization}
 that the testing set
 is independent of the data sequence 
  used for SGM updates in~\eqref{sgm dynamics}. 
\begin{prop}
\label{gen error prop independent}
 Suppose that Assumption~\ref{basic assumptions} holds and
 that the testing set
$$
 Z := (Z_i)_{1 \leq i \leq n}
= (X_i, Y_i)_{1 \leq i \leq n}
\overset{\text{i.i.d.}}{\sim} \rho
$$
 is independent of the training sequence $(Z^{(t)})_{0 \leq t \leq T}$. 
\begin{enumerate}[a)]
\item If $W(t)$ remains frozen at
 $W(t):=w \in \real^{d \times d_{\rm out}}$, 
 i.e. $\eta_W (t) = 0$, $0 \leq t < T$, 
 then we have 
    \begin{equation}
      \label{fjdksf independent}
        \E \big[ \big| \varepsilon_{\rm gen}(n, V(T), w) \big| \big]
        \leq
        (1 + C_1 ( w, T ) ) \frac{2}{\sqrt{n}},
    \end{equation}
    where
$$
    C_1( w, T)
    :=     
        \norm{w}_F 
        \sqrt{\kappa d_{\rm in}}
        \prod\limits_{t = 0}^{T - 1} (1 - \eta (t) \lambda) 
    +
    \frac{\norm{w}_F^2}{\lambda}
    \left(1 - \prod\limits_{t = 0}^{T - 1} (1 - \eta (t) \lambda)\right). 
$$
\item
  If $\eta_W (t) = \eta_V (t):=\eta (t)$, $0 \leq t < T$, 
  then
we have
    \begin{equation}
      \label{fjdksf independent2}
        \E \big[ \big| \varepsilon_{\rm gen}(n, V(T), W(T))\big| \big]
        \leq
        (1 + C_2( 1, T) ) \frac{2}{\sqrt{n}}, 
    \end{equation}
    where $C_2( 1 , T)$ is defined from 
    \begin{equation}
      \label{jkls11a} 
    C_2( p , T) :=
    (2p - 1)!!2^{p-1} 
    \big(d_{\rm in}^p + d^p\big)
    \kappa^p
    \prod\limits_{t = 0}^{T - 1}
            (1 + ( 1 - \lambda ) \eta (t) )^{2p}, 
 \quad p \geq 1. 
\end{equation}
\end{enumerate}
\end{prop}
\begin{Proof}
$a)$ Due to the relations 
$$
    \left\{
    \begin{array}{l}
      \nabla_x l( f ( x , v, w ) , y )
    =  v \Sigma w (\nabla_{y_1} l) ( f ( x , v, w ) , y ), 
    \bigskip
    \\
    \nabla_y l( f ( x , v, w ) , y )
    = (\nabla_{y_2} l) ( f ( x , v, w ) , y ), 
    \end{array}
    \right.
$$ 
 the function $(x,y)\mapsto l( f ( x , v, w ) , y )$ is
    $\left(\norm{v}_F \norm{w}_F\right)$-Lipschitz in $x$
 and $1$-Lipschitz in $y$,
 with 
 \begin{equation}
   \label{jkld19} 
       \big| 
            l( f ( x' , v, w ), y')
            -
            l( f ( x , v, w ), y )
         \big| 
        \leq \norm{v}_F \norm{w}_F |x- x' | + | y'-y|, 
            \end{equation}
        $x,x'\in \real^{d_{\rm in}}$, 
        $y,y'\in \real^{d_{\rm out}}$, 
        $v \in \real^{d_{\rm in} \times d}$,
        $w \in \real^{d \times d_{\rm out}}$. 
 Hence, using H\"older's inequality,
    the independence of $(X_i, Y_i)_{1 \leq i \leq n}$
    and $V(T)$, and the facts
    that for all $z, z' \in \Supp(\rho)$, $|z - z'| \leq 2$
    and \eqref{all bounds}, we have
    \begin{align*}
        & \big| \varepsilon_{\rm gen}(n, V(T), w) \big| 
  = \left\lvert \frac{1}{n} \sum\limits_{i=1}^n
            l(f ( X_i , V(T), w ), Y_i )
        - \int_{\real^{d_{\rm in}}\times \real^{d_{\rm out}}}
            l( f ( x , V(T), w ), y )
            \rho(dx, dy)\right\rvert 
      \\
       & \quad = \left\lvert 
         \int_{\real^{d_{\rm in}}\times \real^{d_{\rm out}}}
            l( f ( x , V(T), w ), y )
(            \tilde{\rho} (dx, dy)
            -
            \rho(dx, dy)
            )
            \right\rvert 
        \\
        & \quad \leq \sqrt{\E \left[
        \left(\frac{1}{n} \sum\limits_{i=1}^n
            l( f ( X_i , V(T), w ), Y_i)
        - \int_{\real^{d_{\rm in}}\times \real^{d_{\rm out}}}
            l( f ( x , V(T), w ), y )
            \rho(dx, dy)\right)^2
        \ \! \bigg| \ \! V(T) \right]} 
        \\
        & \quad = \sqrt{\E \left[
            \left(\frac{1}{n}
            \sum\limits_{i=1}^n
            \int_{\real^{d_{\rm in}}\times \real^{d_{\rm out}}}
            (
            l( f ( X_i , V(T), w ), Y_i)
            -
            l( f ( x , V(T), w ), y )
            )
            \rho(dx, dy)\right)^2
        \ \! \bigg| \ \! V(T) \right]} 
        \\
        & \quad \leq \sqrt{\E \left[
            \frac{1}{n^2}
            \sum\limits_{i=1}^n
            \int_{\real^{d_{\rm in}}\times \real^{d_{\rm out}}}
            (
            l( f ( X_i , V(T), w ), Y_i)
            -
            l( f ( x , V(T), w ), y )
            )^2
            \rho(dx, dy)
        \ \! \bigg| \ \! V(T) \right]} 
        \\
        & \quad \leq \sqrt{
        \frac{4}{n^2} \sum\limits_{i=1}^n \int_{\real^{d_{\rm in}}\times \real^{d_{\rm out}}}
             ( 1 + \norm{V(T)}_F \norm{w}_F )^2 \rho(dx, dy)
        } 
        \\
        & \quad = \frac{2}{\sqrt{n}}
              ( 1 + \norm{V(T)}_F \norm{w}_F )
       ,
    \end{align*}
 where we applied \eqref{jkld19}, hence 
        \begin{align*}
      \E\big[ \big| \varepsilon_{\rm gen}(n, V(T), w) \big| \big]
       \leq
     \frac{2}{\sqrt{n}} ( 1 +
  \norm{w}_F        \E \left[ \norm{V(T)}_F \right] ),
    \end{align*}
        and we conclude by the application of Proposition~\ref{main prop}-$(a)$
        with $p=1$.
 
\noindent
$b)$ By the same argument as in part~$(a)$
we have
$$
 \big| \varepsilon_{\rm gen}(n, V(T), W(T)) \big| 
        \leq
     \frac{2}{\sqrt{n}} ( 1 +
     \E \left[ \norm{W(T)}_F \norm{V(T)}_F \right] ),
     $$ 
      and we conclude by the application of Proposition~\ref{main prop}-$(b)$
      with $p=1$.
\end{Proof}
In Proposition~\ref{gen error prop independent-2},
we present related deviation inequalities. 
\begin{prop}
    \label{gen error prop independent-2}
 Suppose that Assumption~\ref{basic assumptions} holds and
 that the testing set
$$
 Z := (Z_i)_{1 \leq i \leq n}
= (X_i, Y_i)_{1 \leq i \leq n}
\overset{\text{i.i.d.}}{\sim} \rho
$$
 is independent of the training sequence $(Z^{(t)})_{0 \leq t \leq T}$. 
\begin{enumerate}[a)]
\item If $W(t)$ remains frozen at
 $W(t):=w \in \real^{d \times d_{\rm out}}$, 
 i.e. $\eta_W (t) = 0$, $0 \leq t < T$, 
 then for any $\zeta \in (0,1)$ we have 
\begin{equation*}
        \P\left( \big| \varepsilon_{\rm gen}(n, V(T), w)\big| 
            \leq 
            (1 + C_3 ( w, \zeta , T) )
                \sqrt{\frac{2}{n} \log \frac{4}{\zeta}} \ \right)
        \geq 1 - \zeta,
    \end{equation*}
where      $$
    C_3 ( w, \zeta , T): 
=      \norm{w}_F 
\left(
\!\!
\sqrt{d_{\rm in}} + \sqrt{d}
+ \sqrt{2 \log \frac{4}{\zeta}}
\ 
\right)
                    \sqrt{\frac{\kappa}{d}}
                    \prod\limits_{t = 0}^{T - 1} (1 - \eta (t) \lambda)
    +
    \frac{\norm{w}_F^2}{\lambda}
    \left(1 - \prod\limits_{t = 0}^{T - 1} (1 - \eta (t) \lambda)\right).
    $$
  \item
  If $\eta_W (t) = \eta_V (t):=\eta (t)$, $0 \leq t < T$, 
  then for any $\zeta \in (0,1)$ we have 
    \begin{equation*}
        \P\left( \big| \varepsilon_{\rm gen}(n, V(T), W(T))\big| 
            \leq 
            (1 + C_5 ( \zeta , T) )
                \sqrt{\frac{2}{n} \log \frac{4}{\zeta}} \ \right)
        \geq 1 - \zeta,
    \end{equation*}
    where
      $$ C_5 ( \zeta , T) := 
    3 \kappa
    \left( 2 + \frac{d_{\rm in}}{d} + \frac{d}{d_{\rm out}}
        +  \left(\frac{2}{d} + \frac{2}{d_{\rm out}}\right)
        \log \frac{8}{\zeta}\right)
    \prod\limits_{t = 0}^{T - 1} (1 + ( 1 - \lambda ) \eta (t) )^2. 
    $$
\end{enumerate}
\end{prop}
\begin{Proof}
  $a)$
  From \eqref{jkld19}, we have the bound 
    \begin{align*}
        |l(f ( x , v, w ), y)|
       & \leq |l(f ( x , v, w ), y) - l(y, y)| + l(y, y)
        \\
        &
        \leq |f ( x , v, w ) - y|
        \\
        &
        \leq 1 + \norm{v}_F \norm{w}_F, 
            \end{align*}
            $x\in \real^{d_{\rm in}}$, 
        $y\in \real^{d_{\rm out}}$, 
        $v \in \real^{d_{\rm in} \times d}$,
    $w \in \real^{d \times d_{\rm out}}$,
    hence by Hoeffding's inequality, see Theorem~1 in \cite{hoeffding2},
 the generalization error 
    \begin{equation*}
        \varepsilon_{\rm gen}(n, V(T), w) =
              \frac{1}{n} \sum\limits_{i=1}^n
              l(f ( X_i , V(T), w ), Y_i )
              -
                      \int_{\real^{d_{\rm in}}\times \real^{d_{\rm out}}} l(f ( x , V(T), w ), y) \rho(dx, dy)
    \end{equation*}
 satisfies 
    \begin{equation*}
      \P \left( \big| \varepsilon_{\rm gen}(n, V(T), w) \big|
      \leq ( 1 + \norm{V(T)}_F \norm{w}_F )
                \sqrt{\frac{2}{n} \log \frac{2}{\zeta}}
                                \ \ \! \bigg| \ \ \!
                                V(T) \right)
        \geq 1 - \zeta,
    \end{equation*}
    which %
    yields
    \begin{equation}
        \label{hoeffding bound}
        \P \left( \big| \varepsilon_{\rm gen}(n, V(T), w) \big|
        \leq
        ( 1 + \norm{V(T)}_F \norm{w}_F )
        \sqrt{\frac{2}{n} \log \frac{2}{\zeta}}
        \ \right)
        \geq 1 - \zeta.
    \end{equation}
    Next, by \eqref{jklsd18} and
 the bound (2.3) in \cite{rudelson2010non}, which implies 
    \begin{equation}
        \label{random matrix spectral norm}
        \P \left( \norm{V(0)}_F
        \leq
        \sqrt{\frac{\kappa}{d}}
        \left(\sqrt{d_{\rm in}} + \sqrt{d}
        + \sqrt{2\log \frac{4}{\zeta} } \ \right)
        \right)
        \geq 1 - \frac{\zeta}{2}, 
        \quad
        \zeta \in (0, 1), 
    \end{equation}
 we get 
    \begin{equation}
        \label{hfjkd1} 
 \P ( \norm{V(T)}_F \norm{w}_F \leq C_3 ( w, \zeta , T) )
 \geq 1 - \frac{\zeta}{2}. 
\end{equation} 
    Hence, from \eqref{hoeffding bound}-\eqref{hfjkd1}
    and the inequality 
    $\P(A \cap B) \geq \P(A) + \P(B) - 1$,
    we have
    \begin{align}
      \nonumber 
 &       1 - \zeta
      \\
      & 
       \leq \P\left( \big| \varepsilon_{\rm gen}(n, V(T), w) \big|
        \leq
        \sqrt{\frac{2}{n} \log \frac{4}{\zeta}}
        ( 1 + \norm{V(T)}_F \norm{w}_F )
        \right)
      \\
      & 
      \quad
      -1
      + \P\left( \norm{V(T)}_F \norm{w}_F \leq C_3 ( w, \zeta , T) \right) 
        \\
      \nonumber 
        & \leq \P\left( \big| \varepsilon_{\rm gen}(n, V(T), w) \big|
        \leq
        ( 1 + \norm{V(T)}_F \norm{w}_F )
        \sqrt{\frac{2}{n} \log \frac{4}{\zeta}}
            \ \text{ and } \
                \norm{V(T)}_F \norm{w}_F \leq C_3 ( w, \zeta , T) \right)
        \\
        \label{c33}
         & \leq \P\left( \big| \varepsilon_{\rm gen}(n, V(T), w) \big|
        \leq 
        (1 + C_3 ( w, \zeta , T) )
        \sqrt{\frac{2}{n} \log \frac{4}{\zeta}}
        \right)
    \end{align}
    which completes the proof.

    \noindent
    $b)$
    From \eqref{random matrix spectral norm}, we have  
    \begin{align}
      \nonumber 
        1 - \zeta & \leq \left(1 - \frac{\zeta}{2}\right)^2
        \\
      \nonumber 
        & \leq \P \left( \norm{V(0)}_F
        \leq
         \sqrt{\frac{\kappa}{d}}
         \left(\sqrt{d_{\rm in}} + \sqrt{d}
         + \sqrt{2\log \frac{4}{\zeta}} \ \right)
         \right)
        \\
      \nonumber 
        & \qquad \qquad \qquad \qquad \qquad \qquad
            \times \P \left( \norm{W(0)}_F
            \leq
            \sqrt{\frac{\kappa}{d_{\rm out}}}
            \left(\sqrt{d} + \sqrt{d_{\rm out}}
            + \sqrt{2\log \frac{4}{\zeta}} \ \right)
            \right)
        \\
      \nonumber 
        & \leq \P \left( \norm{V(0)}^2_F
            \leq 3 \kappa \left( 1 + \frac{d_{\rm in}}{d} + \frac{2}{d} \log \frac{4}{\zeta} \right)
            \text{ and }
            \norm{W(0)}^2_F
            \leq 3 \kappa \left( 1 + \frac{d}{d_{\rm out}}
        + \frac{2}{d_{\rm out}} \log \frac{4}{\zeta} \right) \right)
        \\
\label{c44} 
        & \leq \P \left( \norm{V(0)}^2_F + \norm{W(0)}^2_F
            \leq 3 \kappa \left( 2 + \frac{d_{\rm in}}{d} + \frac{d}{d_{\rm out}}
                + \left(\frac{2}{d} + \frac{2}{d_{\rm out}}\right)
                \log \frac{4}{\zeta} \right) \right),
    \end{align}
    where the third inequality uses H\"older's inequality and
    the independence of $V(0)$, $W(0)$. 
    We conclude from \eqref{b22} using the same argument as in~\eqref{c33}. 
\end{Proof}

\section{%
Random subset} %
\label{sec:application to gen error}
 In this section, we make no independence assumption between
 the testing set
$$
 Z := (Z_i)_{1 \leq i \leq n}
= (X_i, Y_i)_{1 \leq i \leq n}
\overset{\text{i.i.d.}}{\sim} \rho
$$
and the training sequence
 $(Z^{(t)})_{0\leq t \leq T}$ used for SGM updates in~\eqref{sgm dynamics}. 
In Proposition~\ref{gen error prop}, using Propositions~\ref{fg} and
\ref{main prop}
 we derive bounds on the generalization error 
 \eqref{eq:gen error}
 of \eqref{sgm dynamics} 
 under the technical condition $d_{\rm in} + d_{\rm out} \geq 5$
 which originates in Proposition~\ref{fg}, 
 and can be removed at the expense of additional
 analysis. 
\begin{prop}
\label{gen error prop}
Suppose that $d_{\rm in} + d_{\rm out} \geq 5$
and that Assumption~\ref{basic assumptions}
 holds.
\begin{enumerate}[a)]
   \item Assume that $\eta_W (t) = 0$, $0 \leq t < T$, 
 i.e.       $W(t)$ remains frozen at
 $W(t):=w \in \real^{d \times d_{\rm out}}$,
 $0 \leq t \leq T$. 
 Then, we have
\begin{equation}
\nonumber %
        \E \big[ \big|\varepsilon_{\rm gen}(n, V(T), w)\big| \big]
        \leq
        \frac{
          \sqrt{ ( 1 + C_4 ( w, T) )C}}{
          n^{1 / ( d_{\rm in}+d_{\rm out} ) }}, \quad n \geq 1, 
    \end{equation}
 where
$$
    C_4 ( w, T)
    :=     
    2 \norm{w}_F 
        \kappa d_{\rm in}
        \left(\prod\limits_{t = 0}^{T - 1} (1 - \eta (t) \lambda)\right)^2
    +
   2 \frac{\norm{w}_F^3}{\lambda^2}
    \left(1 - \prod\limits_{t = 0}^{T - 1} (1 - \eta (t) \lambda)\right)^2, 
$$
 and $C>0$ is the constant given in~\eqref{fournier guillin}. 
\item
  If $\eta_W (t) = \eta_V (t):=\eta (t)$, $0 \leq t < T$, 
    then we have
  \begin{equation}
\nonumber %
        \E \big[ \big|\varepsilon_{\rm gen}(n, V(T), W(T))\big| \big]
        \leq
        \frac{
          \sqrt{ ( 1 + C_2( 2, T) )C}}{
          n^{1 / ( d_{\rm in}+d_{\rm out} ) }}
        , \quad n \geq 1, 
    \end{equation}
 where $C_2( 2, T)$ is 
 defined in~\eqref{jkls11a} %
 and $C>0$ is the constant given in~\eqref{fournier guillin}. 
\end{enumerate}
\end{prop}
\begin{Proof}
$a)$ Since  from \eqref{jkld19} the function 
    $
    (x,y ) \mapsto l( f ( x , V(T), w ) , y )
    $
    is
    $\left(\norm{w}_F\norm{V(T)}_F \right)$-Lipschitz in $x$
    and $1$-Lipschitz in $y$, 
 using
    H\"older's inequality, \eqref{kantorovich-rubinstein} 
    and \eqref{all bounds}, we have
    \begin{align}
      \nonumber
   \big|\varepsilon_{\rm gen}(n, V(T), w)\big| 
        & 
       = \left\lvert \frac{1}{n} \sum\limits_{i=1}^n
            l( f ( X_i , V(T), w ), Y_i )
        - \int_{\real^{d_{\rm in}}\times \real^{d_{\rm out}}}
            l( f ( x , V(T), w ), y )
            \rho(dx, dy)\right\rvert 
      \\
\nonumber
            & \quad
      = \left\lvert
        \int_{\real^{d_{\rm in}}\times \real^{d_{\rm out}}}
            l( f ( x , V(T), w ), y )
            (
            \tilde{\rho} (dx, dy)
 - 
            \rho(dx, dy)
) 
            \right\rvert 
        \\
        \label{a1}
         & \quad \leq 
            \mathcal{W}_1(\rho, \tilde{\rho}_n)
            \sqrt{1 + \norm{V(T)}^2_F \norm{w}^2_F }
        ,
    \end{align}
 hence 
        \begin{align*}
        \E\big[ \big|\varepsilon_{\rm gen}(n, V(T), w)\big| \big]
        & \leq \E\left[
            \mathcal{W}_1(\rho, \tilde{\rho}_n)
            \sqrt{1 + \norm{V(T)}^2_F \norm{w}^2_F }
            \right]
        \\
        & \quad \leq \sqrt{
                     \big(
            1 + \norm{w}^2_F \E\left[\norm{V(T)}^2_F \right]
            \big)
            \E\left[\mathcal{W}^2_F (\rho, \tilde{\rho}_n)\right]
        },
    \end{align*}
        which completes the proof by \eqref{fournier guillin}
        and Proposition~\ref{main prop}-$(a)$. 

    \noindent
    $b)$ The argument is the same as in
 part~$(a)$,
    replacing the use of Proposition~\ref{main prop}-$(a)$
    with that of Proposition~\ref{main prop}-$(b)$. 
\end{Proof}
 Similarly, using Proposition~\ref{fg}
 we obtain the following concentration inequality. 
\begin{prop}
 Suppose that $d_{\rm in} + d_{\rm out} \geq 5$
 and that Assumption~\ref{basic assumptions} holds. 
\begin{enumerate}[a)]
\item Assume that $\eta_W (t) = 0$, $0 \leq t < T$, 
 i.e.       $W(t)$ remains frozen at
 $W(t):=w \in \real^{d \times d_{\rm out}}$,
 $0 \leq t \leq T$. Then, for any $\zeta \in (0,1)$ we have
    \begin{equation*}
        \P\left( \big|\varepsilon_{\rm gen}(n, V(T), w)\big|
            \leq 
            \left(
            \frac{Cn}{\log(2C / \zeta)}
            \right)^{- 1 / ( d_{\rm in}+d_{\rm out} ) }
            \! \! \! \! \! 
            (1 + C_3 ( w, \zeta , T) ) \right)
        \geq 1 - \zeta, \quad n \geq 1, 
    \end{equation*}
where
    $C_3 ( \zeta , T)$
    is defined in Proposition~\ref{gen error prop independent-2}-$(a)$
     and $C$ is given in~\eqref{fournier guillin_2}.
\item
  If $\eta_W (t) = \eta_V (t):=\eta (t)$, $0 \leq t < T$, 
  then for any $\zeta \in (0,1)$ we have
    \begin{equation*}
        \P\left( \big|\varepsilon_{\rm gen}(n, V(T), W(T))\big|
            \leq 
            \left(
            \frac{Cn}{\log(2C / \zeta)}
            \right)^{- 1 / ( d_{\rm in}+d_{\rm out} ) }
            \! \! \! \! \! 
                (1 + C_5 ( \zeta , T) ) \right)
        \geq 1 - \zeta, \quad n \geq 1, 
    \end{equation*}
    where
    $C_5 ( \zeta , T)$
    is defined in Proposition~\ref{gen error prop independent-2}-$(b)$. 
\end{enumerate}
\end{prop}
\begin{Proof}
$a)$
 By \eqref{jklsd18} and \eqref{random matrix spectral norm}
 we have 
$$
 \P\big( \norm{V(T)}_F \norm{w}_F \leq C_3 ( w, \zeta , T) \big)
 \geq 1 - \frac{\zeta}{2},
$$
 hence, using the inequality $\P(A \cap B) \geq \P(A) + \P(B) - 1$,
 the bounds \eqref{fournier guillin_2} and
  \eqref{a1}, %
 we have
    \begin{align*}
 &
        1 - \zeta
        \leq \P\left( \mathcal{W}_1(\rho, \tilde{\rho}_n) \leq 
        \left( 
        \frac{Cn}{\log(2C / \zeta)}
        \right)^{- 1/( d_{\rm in}+d_{\rm out} ) } \right)
            + \P\big( \norm{V(T)}_F \norm{w}_F \leq C_3 ( w, \zeta , T) \big) - 1
      \\
        &\leq  \P\left( \mathcal{W}_1(\rho, \tilde{\rho}_n) \leq 
        \left(
        \frac{Cn}{\log(2C / \zeta)}
        \right)^{- 1 / ( d_{\rm in}+d_{\rm out} ) }
        \! \! \! \! \! \text{ and } \ 
                \norm{V(T)}_F \norm{w}_F \leq C_3 ( w, \zeta , T) \right)
        \\
        & \leq  \P\left( \big|\varepsilon_{\rm gen}(n, V(T), w)\big|
            \leq 
            \mathcal{W}_1(\rho , \tilde{\rho}_n)
            \sqrt{1 + \norm{w}^2_F\norm{V(T)}^2_F }
            \leq 
            \left(
            \frac{Cn}{\log(2C / \zeta)}
            \right)^{- 1 / ( d_{\rm in}+d_{\rm out} ) }
            \! \! \! \! \! \!
            \! \! \! \! \! \!
            \! \! \! \! \! \!
            \! \! \! \! \! \!
            (1 + C_3 ( w, \zeta , T)) \right).
    \end{align*}

    \noindent
    $b)$ The proof proceeds as in part~$(a)$, by 
 replacing the uses of \eqref{jklsd18} and 
 \eqref{random matrix spectral norm}
 with those of \eqref{b22} and 
 \eqref{c44}. 
\end{Proof}
 Proposition~\ref{main prop full} presents $L^p$
 bounds on the Lipschitz constant of the loss function
 $\ell_\lambda ( f ( x , v, w ), y , v, w)$. 
\begin{prop}
\label{main prop full}
    Lipschitz bound.
    Suppose that Assumption~\ref{basic assumptions} holds. 
 \begin{enumerate}[a)]
    \item Assume that $\eta_W (t) = 0$, $0 \leq t < T$, 
 i.e.       $W(t)$ remains frozen at
 $W(t):=w \in \real^{d \times d_{\rm out}}$,
 $0 \leq t \leq T$. 
 For $p=1$ we have the gradient bound 
\begin{align*}
    & \E \left[ \sup\limits_{(x, y) \ \! \in \ \! \Supp (\rho)
          }
            |\nabla_x \ell_\lambda ( f ( x , V(T), w ), y , V(T), w)| \right]
       \\
    & \qquad \leq 
    \norm{w}_F 
        \big(\sqrt{d_{\rm in}} + \sqrt{d}\big)
        \sqrt{\frac{\kappa}{d}}
        \left(\prod\limits_{t = 0}^{T - 1} (1 - \eta (t) \lambda)\right)
    +
    \frac{\norm{w}_F^2}{\lambda}
    \left(1 - \prod\limits_{t = 0}^{T - 1} (1 - \eta (t) \lambda)\right), 
\end{align*}
 and for $p \geq 2$ we have 
\begin{align*}
    & \E \left[ \sup\limits_{(x, y) \ \! \in \ \! \Supp (\rho)
          }
            |\nabla_x \ell_\lambda ( f ( x , V(T), w ), y , V(T), w)|^p \right]
    \\
    & \qquad \leq 
        (p - 1)!! 2^{p-1} \norm{w}_F^p (\kappa d_{\rm in})^{p/2}
    \left(\prod\limits_{t = 0}^{T - 1} (1 - \eta (t) \lambda)\right)^p
    +
    2^{p-1} \frac{\norm{w}_F^{2p}}{\lambda^p}
    \left(1 - \prod\limits_{t = 0}^{T - 1} (1 - \eta (t) \lambda)\right)^p. 
\end{align*}
\item
 If $\eta_W (t) = \eta_V (t):=\eta (t)$, $0 \leq t < T$, 
 then for any $p\geq 1$ we have
\end{enumerate}
\begin{align*}
    &  \E \left[ \sup\limits_{(x, y) \ \in \ \Supp (\rho)}
            |\nabla_x \ell_\lambda ( f ( x , V(T), W(T) ), y , V(T), W(T))|^p \right]
      \\
      & \qquad \qquad \qquad
      \qquad \qquad \qquad
       \qquad \qquad \qquad
      \leq 
            \frac{(2p - 1)!!}{2^{1-p}} 
            \big(d_{\rm in}^p +  d_{\rm out}^p \big)
            \kappa^p
            \prod\limits_{t = 0}^{T - 1}
            (1 + ( 1 - \lambda ) \eta (t) )^{2p}.
\end{align*}
\end{prop}
\begin{Proof}
$a)$
 From the relation 
\begin{equation} 
\label{jfkld124}
 \nabla_x \ell_\lambda ( f(x,v,w), y , v, w)
 = v \Sigma w (\nabla_{y_1} l)
 ( f(x,v,w), y )
\end{equation} 
 combined with \eqref{all bounds} and \eqref{jklsd18}, we have
\begin{align}
\label{fjkld13}
      & \sup\limits_{(x, y) \ \in \ \Supp (\rho)
} |\nabla_x \ell_\lambda ( f ( x , V(T), w ), y , V(T), w)|^p
        \leq \norm{w}^p_F \norm{V(T)}^p_F 
        \\
\nonumber 
        & \qquad \qquad \leq 
        2^{p-1}\norm{w}^p_F\norm{V(0)}^p_F
        \left(\prod\limits_{t = 0}^{T-1} (1 - \eta (t) \lambda)\right)^p
        + 2^{p-1} \norm{w}^{2p}_F \lambda^{-p}
        \left(1 - \prod\limits_{t = 0}^{T - 1} (1 - \eta (t) \lambda)\right)^p,
\end{align}
 and we conclude by the bound \eqref{jkls13-0}. %

\noindent
$b)$ 
 Using H\"older's inequality, the bound \eqref{all bounds} and 
 \eqref{jfkld124}, as in~\eqref{b22} %
 we have
\begin{align}
      \label{fjkld15}
    & \sup\limits_{(x, y) \ \in \ \Supp (\rho)
}
 |\nabla_x \ell_\lambda ( f ( x , V(T), W(T) ), y , V(T), W(T))|^p
            \leq \norm{V(T)}^p_F \norm{W(T)}^p_F
            \\
            \nonumber
            & \qquad \qquad \qquad
             \qquad \qquad \qquad
             \qquad \qquad
            \leq 2^{p-1}
            \left(\norm{V(0)}^{2p}_F + \norm{W(0)}_F^{2p}\right)
            \prod\limits_{t = 0}^{T - 1}
                (1 + ( 1 - \lambda ) \eta (t))^{2p},
    \end{align}
 and we conclude similarly to part~$(a)$. 
\end{Proof}
 As a consequence of the above arguments,
 we have the following concentration inequality. 
\begin{corollary}
   \label{fjkl132}
    Suppose that Assumption~\ref{basic assumptions} holds. 
 \begin{enumerate}[a)]
    \item Assume that $\eta_W (t) = 0$, $0 \leq t < T$, 
 i.e.       $W(t)$ remains frozen at
 $W(t):=w \in \real^{d \times d_{\rm out}}$,
 $0 \leq t \leq T$. 
 For any $\zeta \in (0,1)$, 
 with probability at least $1 - \zeta$ we have
 the concentration inequality 
\begin{align*}
    & \sup\limits_{(x, y) \ \! \in \ \! \Supp (\rho)
          }
            |\nabla_x \ell_\lambda ( f ( x , V(T), w ), y , V(T), w)|
    \\
    & \qquad \leq 
      \norm{w}_F 
      \left(
      \!\!
      \sqrt{d_{\rm in}} + \sqrt{d}
      + \sqrt{2 \log \frac{2}{\zeta}}
      \ 
      \right)
                    \sqrt{\frac{\kappa}{d}}
                    \prod\limits_{t = 0}^{T - 1} (1 - \eta (t) \lambda)
    +
    \frac{\norm{w}_F^2}{\lambda}
    \left(1 - \prod\limits_{t = 0}^{T - 1} (1 - \eta (t) \lambda)\right). 
   \end{align*}
\item
   If $\eta_W (t) = \eta_V (t):=\eta (t)$, $0 \leq t < T$, 
   then for any $\zeta \in (0,1)$,
   with probability at least $1 - \zeta$ we have
       the concentration inequality
       \begin{align*}
    & \sup\limits_{(x, y) \ \in \ \Supp (\rho)
    }
    |\nabla_x \ell_\lambda ( f ( x , V(T), W(T) ), y , V(T), W(T))|
    \\
    &
    \qquad \qquad \qquad
    \qquad \qquad 
    \leq 
    3 \kappa
    \left( 2 + \frac{d_{\rm in}}{d} + \frac{d}{d_{\rm out}}
        + \left(\frac{2}{d} + \frac{2}{d_{\rm out}}\right)
        \log \frac{4}{\zeta}\right)
    \prod\limits_{t = 0}^{T - 1} (1 + ( 1 - \lambda ) \eta (t) )^2. 
\end{align*}
   \end{enumerate}
\end{corollary}
\begin{Proof}
\noindent $a)$
 This inequality follows from the bounds \eqref{jklsd18},
 \eqref{random matrix spectral norm} and \eqref{fjkld13}.
 
\noindent $b)$
 This inequality follows from the bounds \eqref{b22},
 \eqref{c44} and \eqref{fjkld15}.
\end{Proof} 

\section{Numerical results}
\label{sec:numerical results}
In this section we present the numerical simulations
 for the bounds derived
in Propositions~\ref{gen error prop independent}.
For this, we consider
$$
Y = \max\big( \min \big( \beta^\top X + \epsilon, 1 \big), -1 \big),
$$
where $X$ follows a uniform distribution
on the $100$-dimensional unit sphere $S^{99}$,
$\beta$ is a fixed point on $S^{99}$,
and $\epsilon$ follows a standard normal distribution.
Here, $\rho$ is the corresponding distribution of $(X, Y)$.

\medskip

In the He initialization, we use a centered normal distribution
with variance $1/500$ for every entry of the matrix $V(0)$.
Subsequently,
$V(t) \in \real^{100 \times 1000}$ is updated according to \eqref{sgm dynamics},
and $W(t):=w$ is frozen on the $1000$-dimensional unit sphere. %

\medskip

We use the ReLU activation function $\sigma(x) := \max(0, x)$,
the $L^1$ loss function $l(x, y) := |x - y|$,
the regularization parameter $\lambda := 0.1$,
the learning rate $\eta_V (t) = \eta_W (t) := 0.01$,
$T := 300$ epochs, $n := 250, 500, \dots, 5000$
 samples with the batch size $k := n/10$.
The simulation is repeated 20 times, and
 at each time we record the samples of
the absolute generalization error
$\big| \varepsilon_{gen}(n, V(T), w) \big|$.

\medskip

The mean absolute generalization error
in Proposition~\ref{gen error prop independent} is approximated
using the mean absolute value of the recorded samples.
 Since the true loss $\mathcal{L}(V(T), w)$ in~\eqref{lwv}
 is not available in closed form,
 it is approximated by Monte Carlo simulations with sample size $10^5$.

\begin{figure}[H]
  \centering
 \begin{subfigure}[b]{0.5\textwidth}
    \includegraphics[height=5cm,keepaspectratio]{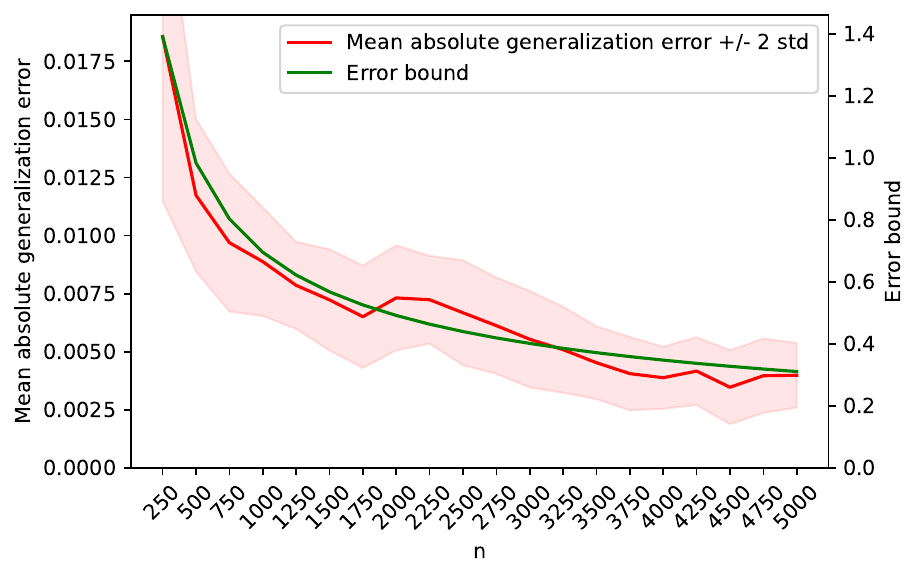}
    \caption{Dual scale.}
 \end{subfigure}
 \begin{subfigure}[b]{0.45\textwidth}
    \includegraphics[height=5cm,keepaspectratio]{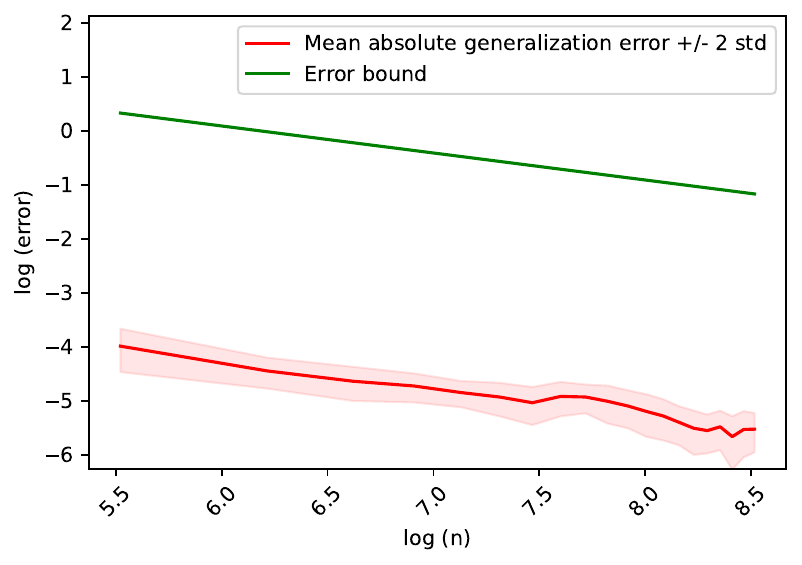}
    \caption{Log scale.}
 \end{subfigure}
  \caption{Mean absolute value of $\varepsilon_{gen}(n, V(T), w)$
          {\em vs.} error bound \eqref{fjdksf independent}.}
\label{fig:comparison}
\end{figure}

\vskip-0.3cm

\noindent
 In Figure~\ref{fig:comparison} we compare
the mean absolute value of the generalization error
to the uniform error bound \eqref{fjdksf independent}
 derived in Proposition~\ref{gen error prop independent}-$a)$.
Figure~\ref{fig:comparison}-$a)$ is plotted on a dual scale by
matching the maximum (initial) values of the two curves to a same level on the graph. 

\medskip

\noindent
In Table~\ref{table:regression} we present the log-log linear regression
displayed in Figure~\ref{fig:comparison}-b),
which confirms the rate of $O(n^{-1/2})$ obtained in Proposition~\ref{gen error prop independent}.

\begin{table}[H]
    \centering
    \begin{tabular}{|c|c|c|c|c|c|}
    \hline
         & Mean     & Stdev    & $t$-statistics & $p$-value  & 95\% conf. interval \\
    \hline
intercept   &  -1.1588 &    0.228 &   -5.094 &    0.000 &    (-1.637, -0.681) \\
\hline
slope &  -0.5139 &    0.030 &  -17.345 &    0.000 &    (-0.576, -0.452) \\
    \hline
    \end{tabular}
    \caption{Log-log regression of the
             mean absolute generalization error.}
    \label{table:regression}
\end{table}

\vskip-0.3cm

\begin{figure}[H]
  \centering
 \begin{subfigure}[b]{0.5\textwidth}
    \includegraphics[height=5cm,keepaspectratio]{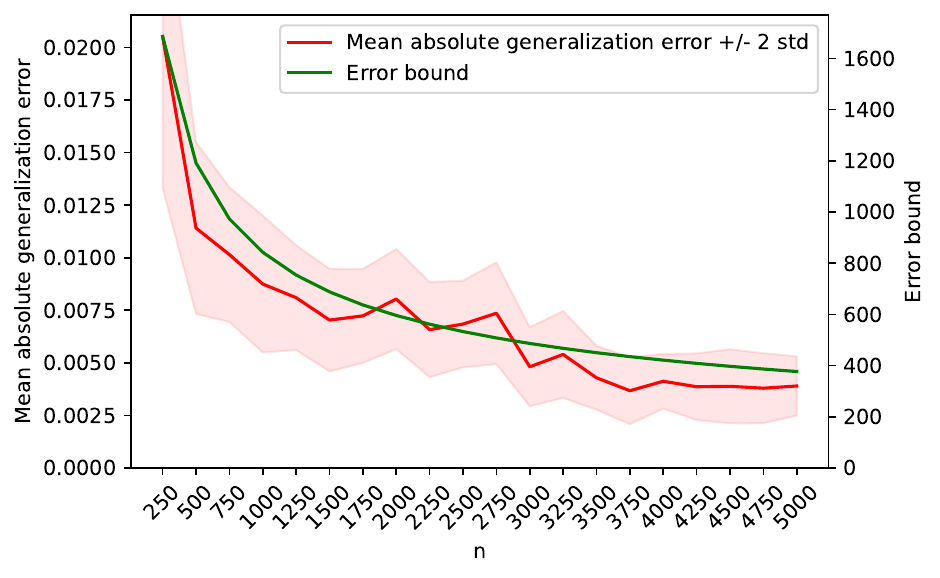}
    \caption{Dual scale.}
 \end{subfigure}
 \begin{subfigure}[b]{0.45\textwidth}
    \includegraphics[height=5cm,keepaspectratio]{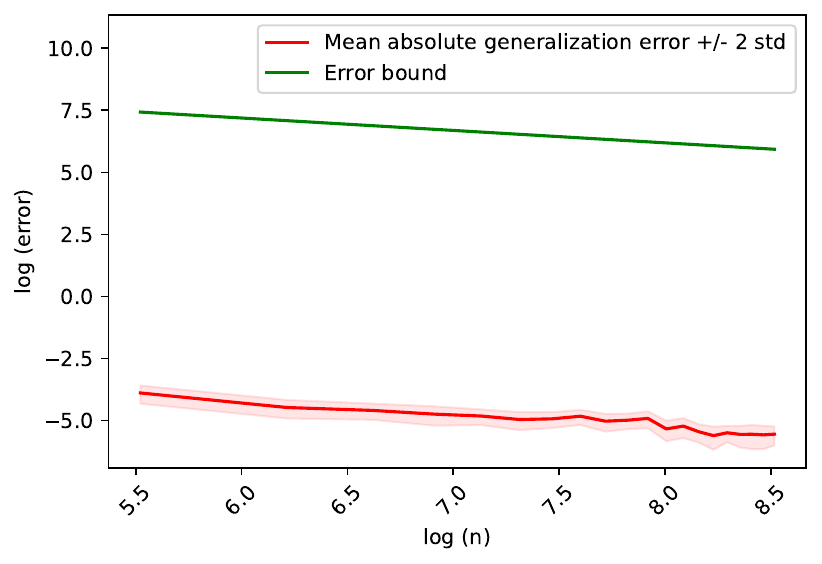}
    \caption{Log scale.}
 \end{subfigure}
 \caption{Mean absolute value of
   $\varepsilon_{gen}(n, V(T), W(T))$ {\em vs.}
  error bound \eqref{fjdksf independent2}.}
\label{fig:comparison unfreezed}
\end{figure}

\noindent
 Next, we run simulations without freezing $W(t)$,
i.e. each element in $W(0)$ follows a centered normal distribution
with variance $2$ and subsequently updated according to \eqref{sgm dynamics}.
In Figure~\ref{fig:comparison unfreezed} we compare
the mean absolute value of the generalization error
and the error bound derived in Proposition~\ref{gen error prop independent}$-b)$ in the same setting as Figure~\ref{fig:comparison}.

\vskip-0.3cm

\noindent
In Table~\ref{table:regression unfreezed} we present the log-log linear regression
displayed in Figure~\ref{fig:comparison unfreezed}-b),
which confirms the rate of $O(n^{-1/2})$
obtained in Proposition~\ref{gen error prop independent}.

\begin{table}[H]
    \centering
    \begin{tabular}{|c|c|c|c|c|c|}
    \hline
         & Mean     & Stdev    & $t$-statistics & $p$-value & 95\% conf. interval \\
    \hline
intercept    &  -0.9745 &    0.292 &   -3.333 &    0.004 &    (-1.589, -0.360) \\
\hline
slope &  -0.5366 &    0.038 &  -14.095 &    0.000 &    (-0.617, -0.457) \\
    \hline
    \end{tabular}
    \caption{Log-log regression of the mean absolute generalization error.}
    \label{table:regression unfreezed}
\end{table}

\vskip-0.3cm

\noindent
We note that although the constants
$C_1(w,T)$, $C_2(2,T)$ in Figures~\ref{fig:comparison}
and \ref{fig:comparison unfreezed}
can be quite large, the $O(n^{-1/2})$ rate of decrease has
been correctly identified.
As the dimension-dependent bounds
in Proposition~\ref{gen error prop} %
are not sharp,
 the numerical simulations based on them are not presented.

\subsubsection*{Acknowledgement}
 We thank the anonymous referees for useful suggestions and
 corrections. 

 \footnotesize

\newcommand{\etalchar}[1]{$^{#1}$}
\def\cprime{$'$} \def\polhk#1{\setbox0=\hbox{#1}{\ooalign{\hidewidth
  \lower1.5ex\hbox{`}\hidewidth\crcr\unhbox0}}}
  \def\polhk#1{\setbox0=\hbox{#1}{\ooalign{\hidewidth
  \lower1.5ex\hbox{`}\hidewidth\crcr\unhbox0}}} \def\cprime{$'$}

\end{document}